\pdfoutput=1
\documentclass{llncs}

\usepackage[T1]{fontenc}
\usepackage[english]{babel}
\usepackage{times}
\usepackage{graphicx}
\usepackage{amsmath}
\usepackage{amsfonts}
\usepackage{wasysym}
\usepackage{url}

\usepackage[boxed,vlined,linesnumbered,ruled]{algorithm2e}

\usepackage{color}
\usepackage{ucs}
\usepackage{subfigure}
\usepackage{latexsym}
\usepackage{amsfonts}
\usepackage{amssymb}
\usepackage{multirow}
\usepackage{multicol}
\usepackage[autolanguage]{numprint}
\usepackage{array}

\usepackage{times}
\usepackage{xcolor,colortbl}
\definecolor{green}{rgb}{0.1,0.1,0.1}

\DeclareMathOperator*{\argmin}{arg\,min}

\usepackage[utf8x]{inputenc}

\begin{document}

\title{A Multistage Stochastic Programming Approach to\\
the Dynamic and Stochastic VRPTW\\
\emph{Extended version}}

\author{Michael Saint-Guillain$^{*}$, Yves Deville$^{*}$ \& Christine Solnon$^{**}$}

\institute{$^{*}$ICTEAM, Université catholique de Louvain, Belgium\\
$^{**}$Université de Lyon, CNRS\\
$^{**}$INSA-Lyon, LIRIS, UMR5205, F-69621, France}
\maketitle
\begin{abstract}
We consider a dynamic vehicle routing problem with time windows and stochastic customers (DS-VRPTW), such that customers may request for services as vehicles have already started their tours. To solve this problem, the goal is to provide a decision rule for choosing, at each time step, the next action to perform in light of known requests and probabilistic knowledge on requests likelihood.
We introduce a new decision rule, called Global Stochastic Assessment (GSA) rule for the DS-VRPTW, and we compare it with existing decision rules, such as MSA. In particular, we show that GSA fully integrates nonanticipativity constraints so that it leads to better decisions in our stochastic context.
We describe a new heuristic approach for efficiently approximating our GSA rule. We introduce a new waiting strategy. Experiments on dynamic and stochastic benchmarks, which include instances of different degrees of dynamism, show that not only our approach is competitive with state-of-the-art methods, but also enables to compute meaningful offline solutions to fully dynamic problems where absolutely no a priori customer request is provided.
\end{abstract}

\section{Introduction\label{sec:Introduction}}

This paper is an extended version of study \cite{saintguillain2015multistage}, and contains additional detailed experimental results.

Dynamic (or \emph{online}) vehicle routing problems (D-VRPs) arise when information about demands is incomplete, {\em e.g.}, whenever a customer is able to submit a request during the online execution of a solution. 
D-VRP instances usually indicate the deterministic requests, {\em i.e.}, those that are known before the online process if any. Whenever some additional (stochastic) knowledge about unknown requests is available, the problem is said to be \emph{stochastic}. 
We focus on the \emph{Dynamic} and \emph{Stochastic} VRP with \emph{Time Windows} (DS-VRPTW). These problems arise in many practical situations, as door-to-door or door-to-hospital transportation of elderly or disabled persons. In many countries, authorities try to set up dial-a-ride services, but escalating operating costs and the complexity of satisfying all customer demands become rapidly unmanageable for solution methods based on human choices \cite{Cordeau2003}. However, such complex dynamic problems need reliable and efficient algorithms that should first be assessed on reference problems, such as the DS-VRPTW.

In this paper, we present \emph{a new heuristic method for solving
the DS-VRPTW}, based on a Stochastic Programming modeling. By definition,
our approach enables a \emph{higher level of anticipation} than heuristic state-of-the-art
methods. The resulting new online decision rule, called Global Stochastic
Assessment (GSA), comes with a theoretical analysis that clearly defines
the nature of the method. We propose a \emph{new waiting strategy}
together with a heuristic algorithm that embeds GSA. We compare GSA with the state-of-the-art method MSA from \cite{bent2004scenario}, and provide a \emph{comprehensive
experimental study} that highlights the contributions of existing and new waiting and relocation strategies.

This paper is organized as follows. Section \ref{sec:problem_desc}
describes the problem. Section \ref{sec:related_work} presents the state-of-the-art method we compare
to and briefly discuss related works. GSA is then presented in Section
\ref{sec:GSA}. Section \ref{sec:Solving-DSVRPTW}
describes an implementation that embeds GSA, based on heuristic local
search. Finally, section \ref{sec:Experimentations} resumes the experimental
results. A conclusion follows in section \ref{sec:Conclusions}.

\section{Description of the DS-VRPTW}\label{sec:problem_desc}

\paragraph{Notations.}
We note $[l,u]$ the set of all integer values $i$ such that $l\leq i\leq u$.
A sequence $<x^i, x^{i+1}, \ldots, x^{i+k}>$ (with $k\geq 0$)  is noted $x^{i..i+k}$, and the concatenation of two sequences $x^{i..j}$ and $x^{j+1..k}$ (with $i\leq j< k$) is noted $x^{i..j}.x^{j+1..k}$. Random variables are noted $\boldsymbol\xi$ and their realizations are noted $\xi$. We note $\xi\in \boldsymbol\xi$ the fact that $\xi$ is a realization of $\boldsymbol\xi$, and $p(\boldsymbol\xi=\xi)$ the probability that the random variable $\boldsymbol\xi$ is realized to $\xi$. Finally, we note $\mathbb{E}_{\boldsymbol\xi} [f(\xi)]$ the expected value of $f(\boldsymbol\xi)$ which is defined by
$\mathbb{E}_{\boldsymbol\xi} [f(\xi)] = \sum_{\xi\in \boldsymbol\xi} p(\boldsymbol\xi=\xi)\cdot f(\xi)$.

\paragraph{Input Data of a DS-VRPTW.}
We consider a discrete time horizon $[1,H]$ such that each online event or decision occurs at a discrete time $t\in [1,H]$,  whereas each offline event or decision occurs at time $t=0$. The DS-VRPTW is  defined on a complete and directed graph $G=(V,E)$. The set of vertices $V=[0,n]$ is composed of a depot (vertex $0$) and $n$ customer regions (vertices $1$ to $n$). To each arc $(i,j)\in E$ is associated a travel time $t_{i,j} \in \mathbb{R}_{\ge 0}$, that is the time needed by a vehicle to travel from $i$ to $j$, with $t_{i,j}\neq t_{j,i}$ in general. To each customer region $i \in [1,n]$ is associated a load $q_i$, a service duration $d_i \in [1,H]$ and a time window $[e_i,l_i]$ with $e_i, l_i \in [1,H]$ and $e_i \le l_i$. 

The set of all customer requests is $R\subseteq [1,n]\times[0,H]$. For each request $(i,t)\in R$, $t$ is the time when the request is revealed. When $t=0$, the request is known before the online execution and it is said to be {\em deterministic}. When $t>0$, the request is revealed during the online execution at time $t$ and it is said to be {\em online} (or dynamic). There may be several requests for a same vertex $i$ which are revealed at different times.
During the online execution, we only know a subset of the requests of $R$ ({\em i.e.}, those which have already been revealed).
However, for each time $t\in [1,H]$, we are provided a probability vector $P^t$ such that, for each vertex $i\in [1,n]$, $P^t[i]$ is the probability that a request is revealed for $i$ at time $t$. 

There are $k$ vehicles and all vehicles have the same capacity $Q$.

\paragraph{Solution of a DS-VRPTW.}
At the end of the time horizon, a solution is a subset of requests $R_a\subseteq R$ together with $k$ routes (one for each vehicle). Requests in $R_a$ are said to be {\em accepted}, whereas requests in $R\setminus R_a$ are said to be {\em rejected}. The routes must satisfy the constraints of the classical VRPTW restricted to the subset $R_a$ of accepted requests, {\em i.e.}, each route must start at the depot at a time $t\geq 1$ and end at the depot at a time $t'\leq H$, and for each accepted request $(i,t)\in R_a$, there must be exactly one route that arrives at vertex $i$ at a time $t’\in [e_i,l_i]$ with a current load $l\leq Q-q_i$ and leaves vertex $i$ at a time $t''\geq t’+d_i$. 
The goal is to minimize the number of rejected requests.

As not all requests are known at time 0, the solution cannot be computed offline, and it is computed during the online execution. More precisely, at each time $t\in [1,H]$, an action $a^t$ is computed.
Each action $a^t$ is composed of two parts: first, for each request $(i,t)\in R$ revealed at time $t$, the action $a^t$ must either accept the request or reject it; second, for each vehicle, the action $a^t$ must give operational decisions for this vehicle at time $t$ ({\em i.e.}, service a request, travel towards a vertex, or wait at its current position).
Before the online execution (at time 0), some decisions are computed offline. Therefore, we also have to compute a first action $a^0$.

A solution is a sequence of actions $a^{0..H}$ which covers the whole time horizon. This sequence must satisfy VRPTW constraints, {\em i.e.}, the actions of $a^{0..H}$ must define $k$ routes such that each request accepted in $a^{0..H}$ is served once by one of these routes within the time window associated with the served vertex and without violating capacity constraints. We define the objective function $\omega$ such that
$\omega(a^{0..t})$ is $+\infty$ if $a^{0..t}$ does not satisfy VRPTW constraints, and $\omega(a^{0..t})$ is the number of requests rejected in $a^{0..t}$ otherwise.
Hence, a solution is a sequence $a^{0..H}$ such that $\omega(a^{0..H})$ is minimal at the end of the horizon.

\paragraph{Stochastic program.}
There are different notations used for formulating stochastic programs; we mainly use those from [8]. 
For each time $t\in [1,H]$, we have a vector of random variables $\boldsymbol\xi^t$ such that, for each vertex $i\in[1,n]$, $\boldsymbol\xi^t[i]$ is realized to $1$ if a request for $i$ is revealed at time $t$, and to $0$ otherwise. The probability distribution of $\boldsymbol\xi^t$ is defined by $P^t$, {\em i.e.}, $p(\boldsymbol\xi^t[i]=1) = P^t[i]$ and $p(\boldsymbol\xi^t[i]=0) = 1-P^t[i]$. We note $\boldsymbol\xi^{1..t}$ the random matrix composed of the random vectors $\boldsymbol\xi^1$ to $\boldsymbol\xi^t$. A realization $\xi^{1..H}\in \boldsymbol\xi^{1..H}$ is called a {\em scenario}.

At each time $t\in [1,H]$, the action $a^t$ must contain one accept or reject for each request which is revealed in $\xi^t$. Therefore, we note $A(\xi^t)$ the set of all actions that contain an accept or a reject for each vertex $i\in [1,n]$ such that $\xi^t[i] = 1$. Of course, these actions also contain other decisions related to the $k$ vehicles. We also note $A(\xi^{t1..t2})$ the sequence of sets $<A(\xi^{t1}), \ldots, A(\xi^{t2})>$ where $t1\leq t2$.

\begin{figure}[t]
\begin{multicols}{2}
\includegraphics[width=.35\textwidth]{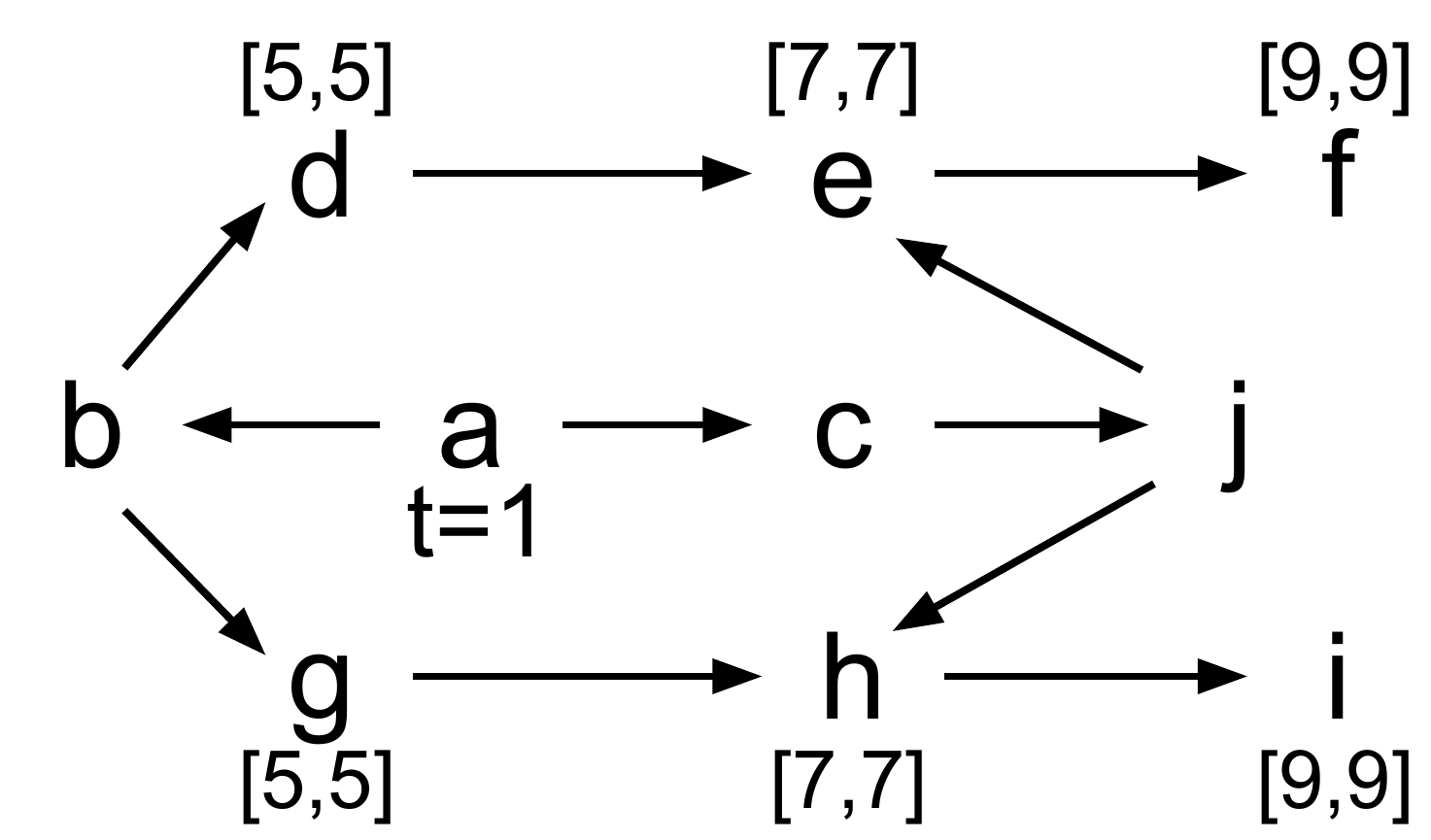}
\columnbreak

\begin{tabular}{|r|c|c|c|c|c|c|c|c|}
\hline
Time & 2 & 3 & 4 & 5 & 6 & 7 & 8 & 9 \\\hline
Scenario $\xi_1^{2..5}$ & $\emptyset$& $\emptyset$ & $\{d,e,f\}$& $\emptyset$& $\emptyset$& $\emptyset$& $\emptyset$& $\emptyset$\\\hline
Scenario $\xi_2^{2..5}$ & $\emptyset$& $\emptyset$& $\{g,h,i\}$& $\emptyset$& $\emptyset$& $\emptyset$& $\emptyset$& $\emptyset$\\\hline
\end{tabular}

~

At time $t=1$, there is only 1 vehicle which is on vertex $a$, and we have to choose between 2 possible actions: {\em travel to b} or {\em travel to c}
\end{multicols}
\protect\caption{A simple example of nonanticipation. The graph is displayed on the left. Time windows are displayed in brackets. For every couple of vertices $(i,j)$, if an arrow $i\rightarrow j$ is displayed then $t_{i,j}=2$; otherwise $t_{i,j}=20$. To simplify, we consider only 2 equiprobable scenarios (displayed on the right). These scenarios have the same prefix (at times $2$ and $3$ no request is revealed) but reveal different requests at time $4$. When using eq. (1) at time $t=1$, we choose to travel to $c$ as the expected cost with nonanticipativity constraints is 1: At time $4$, only one scenario will remain and if this scenario is $\xi_1$ (resp. $\xi_2$), request $(d,4)$ (resp. $(g,4)$) will be rejected. When using eq. (2), we choose to travel to $b$ as the expected cost without nonanticipativity constraints is 0 (for each possible scenario, there exists a sequence of actions which serves all requests: travel to $d$, $e$, and $f$ for $\xi_1$ and travel to $g$, $h$, and $i$ for $\xi_2$). However, if we travel to $b$, at time 3 we will have to choose between traveling to $d$ or $g$ and at this time the expected cost of both actions will be $1.5$: If we travel to $d$ (resp. $g$), the cost with scenario $\xi_1$ is 0 (resp. 3) and the cost with scenario $\xi_2$ is 3 (resp. 0). In this example, the nonanticipativity contraints of multistage problem (\ref{eq:multistage}) thus leads to a better action than the two-stage relaxation (\ref{eq:twostage}).
\label{fig:nonanticipativity_example}}
\end{figure}

Hence, at each time $t$, given the sequence $a^{0..t-1}$ of past actions, the best action $a^t$ is obtained by solving the multistage stochastic problem defined by eq. (\ref{eq:multistage}):
{\small
\begin{equation}
a^t = \hspace{-0.1em} \underset{a^{t}\in A({\xi^t})}{\text{argmin}} \mathbb{E}_{\boldsymbol{\xi}^{t+1}}\left[\min_{a^{t+1}\in A(\xi^{t+1})}\mathbb{E}_{\boldsymbol{\xi}^{t+2}}\left[\cdots\min_{a^{H-1}\in A(\xi^{H-1})}\mathbb{E}_{\boldsymbol{\xi}^{H}}\left[\min_{a^{H}\in A(\xi^{H})} \omega(a^{0..H})\right]\cdots\right]\right]\label{eq:multistage}
\end{equation}
}
Note that this multistage stochastic problem is different from the two-stage stochastic problem defined by eq. (\ref{eq:twostage}):
\begin{equation}
a^t = \argmin_{a^t\in A({\xi^t})} \mathbb{E}_{\boldsymbol\xi^{t+1..H}} [\min_{a^{t+1..H}\in A(\xi^{t+1..H})} \omega(a^{0..H})]
\label{eq:twostage}
\end{equation}
Indeed, eq. (\ref{eq:multistage}) enforces  {\em nonanticipativity constraints} so that, at each time $t'>t$, we consider the action $a^{t'}$ which minimizes the expectation with respect to $\boldsymbol\xi^{t'}$ only, without considering the possible realizations of $\boldsymbol\xi^{t'+1..H}$. Eq. (\ref{eq:twostage}) does not enforce these constraints and considers the best sequence $a^{t+1..H}$ for each realization $\xi^{t+1..H}\in\boldsymbol\xi^{t+1..H}$. 
Therefore, eq. (\ref{eq:multistage}) may lead to a larger expectation of $\omega$ than eq. (\ref{eq:twostage}), as it is more constrained. However, the expectation computed in eq. (\ref{eq:multistage}) leads to better decisions in our context where some requests are not revealed at time $t$. This is illustrated in Fig. \ref{fig:nonanticipativity_example}. 

\section{Related Work}\label{sec:related_work}
The first D-VRP is proposed in \cite{wilson1977computer}, which introduces a single vehicle Dynamic Dial-a-Ride Problem (D-DARP) in which customer requests appear dynamically. Then, \cite{psaraftis1980dynamic} introduced the concept of immediate requests that must be serviced as soon as possible, implying a replanning of the current vehicle route. Complete reviews on D-VRP may be found in \cite{psaraftis1995dynamic,pillac2013review}. In this section, we more particularly focus on stochastic D-VRP.
\cite{pillac2013review} classifies approaches for stochastic D-VRP  in two categories, either based on \emph{stochastic modeling} or on \emph{sampling}.
Stochastic modeling approaches formally capture the stochastic nature of the problem, so that solutions are computed in the light of an overall stochastic context. Such holistic approaches usually require strong assumptions and efficient computation of complex expected values. Sampling approaches try to capture stochastic knowledge by
sampling scenarios, so that they tend to be more focused on local stochastic evidences. Their local decisions however allow sample-based methods to scale up to larger problem instances, even under challenging timing constraints. One usually needs to find a good compromise between having a high number of scenarios, providing a better representation of the real distributions, and a more restricted number of these leading to less computational effort. 

\cite{bent2004scenario} studies the DS-VRPTW and introduces the Multiple Scenario Approach (MSA). A key element of MSA is an adaptive memory that stores a pool of solutions. Each solution is computed by considering a particular scenario which is optimized for a few seconds. The pool is continuously populated and filtered such that all  solutions are consistent with the current system state. Another important element of MSA is the \emph{ranking function} used to make operational decisions involving
idle vehicles. The authors designed 3 algorithms for that purpose:
\begin{itemize}
\item \emph{Expectation} \cite{Bent2004a,Bent2004b} samples a set of scenarios and selects the next request to be serviced by considering its average cost on the sampled set of scenarios. Algorithm \ref{alg:expectation-algo} \cite{VanHentenryck2009} depicts how it chooses the next action $a^t$ to perform. It requires an optimization for each action $a^t\in A(\xi^t)$ and each scenario $s\in S$ (lines 3-4), which is computationally very expensive, even with a heuristic approach.

\item \emph{Regret} \cite{Bent2004a,Bent2005a} approximates the expectation algorithm by recognizing that, given a solution $sol_{s}^{*}$ to a particular scenario $s$, it is possible to compute a good approximation of the local loss inquired by performing another action than the next planned one in $sol_{s}^{*}$.
\item \emph{Consensus }\cite{Bent2004b,bent2004scenario} selects the request that appears the most frequently as the next serviced request in the solution pool.
\end{itemize}
\begin{algorithm}[t]
\lFor{$a^t\in A(\xi^t)$}{
	$f(a^t)\leftarrow{0}$
}
Generate a set $S$ of $\alpha$ scenarios using Monte Carlo sampling\\
	\For{each scenario $s\in S$ and each action $a^t \in A(\xi^t)$}{
		$f(a^t)\leftarrow f(a^t)+$cost of (approximate) solution to scenario $s$ starting with $a^t$
	}
\Return{$\argmin_{a^t \in A(\xi^t)}f(a^t)$}
\protect\caption{The ChooseRequest-$\varepsilon$ Expectation Algorithm\label{alg:expectation-algo}}
\end{algorithm}

Quite similar to the consensus algorithm is the Dynamic Sample Scenario Hedging Heuristic introduced by \cite{Hvattum2006} for the stochastic VRP. 
Also, \cite{Ichoua2006} designed a Tabu Search heuristic for the DS-VRPTW and introduced a vehicle-waiting strategy computed on a future request probability threshold in the near region. Finally, \cite{Bent2007} extends MSA with waiting and relocation strategies so that the vehicles are now able to relocate to promising but unrequested yet vertices. As the performances of MSA has been demonstrated in several studies \cite{Bent2007,flatberg2007dynamic,Schilde2011,Pillac2012}, it is still considered as a state-of-the-art method for dealing with DS-VRPTW.

Other studies of particular interest for our paper are \cite{Ghiani2009}, on the dynamic and stochastic pickup and delivery problem,  and  \cite{Schilde2011}, on the DS-DARP. Both consider local search based algorithms. Instead of a solution pool, they exploit one single solution that minimizes the expected cost over a set of scenarios. However, in
order to limit computational effort, only near future requests are sampled within each scenario. Although the approach of \cite{Schilde2011} is similar to the one of \cite{Ghiani2009}, the set of scenarios considered is reduced to one scenario. Although these later papers show some similarities with the approach we propose, they do not provide any mathematical motivation and analysis of their methods.

\section{The global Stochastic Assessment decision rule} \label{sec:GSA}


The two-stage stochastic problem defined by eq. (\ref{eq:twostage}) may be solved by a sampling solving method such as MSA, which solves a deterministic VRPTW for each possible scenario (i.e., realization of the random variables) and selects the action $a^t$ which minimizes the sum of all minimum objective function values weighted by scenario probabilities. However, we have shown in Section 2 that eq. (2) does not enforce nonanticipativity constraints because the different deterministic VRPTW are solved independently. 
To enforce nonanticipativity constraints while enabling sampling methods, we push these constraints in the computation of the optimal solutions for all different scenarios: Instead of computing these different optimal solutions independently, we propose to compute them all together so that we can ensure that whenever two scenarios share a same prefix of realizations, the corresponding actions are enforced to be equal. 

At each time $t\in[0,H]$, let $r$ be the number of different possible realizations of $\boldsymbol\xi^{t+1..H}$, and let us note $\xi^{t+1..H}_1, \ldots, \xi^{t+1..H}_r$ these realizations. Given the sequence $a^{0..t-1}$ of past actions, we choose action $a^{t}$ by using eq. (\ref{eq:exact_two_stage})
\begin{equation}
a^{t}=\argmin_{a^{t}\in A(\xi^{t})}{\cal Q}(a^{0..t},\{\xi^{t+1..H}_1, \ldots, \xi^{t+1..H}_r\})\label{eq:exact_two_stage}
\end{equation}
which is called the deterministic equivalent form of eq. (\ref{eq:multistage}).\\
${\cal Q}(a^{0..t},\{\xi^{t+1..H}_1, \ldots, \xi^{t+1..H}_r\})$ solves the deterministic optimization problem 
\begin{align}
\min_{a^{t+1..H}_1\!\in A(\xi^{t+1..H}_1), \ldots, a^{t+1..H}_r\!\in A(\xi^{t+1..H}_r)}  ~ \sum_{i=1}^r p(\boldsymbol{\xi}^{t+1..H}\!\!=\!\!\xi_{i}^{t+1..H})\cdot\omega(a^{0..t}.a_{i}^{t+1..H})\label{eq:multistage_deterministic_equivalent}\\
s.t. ~(\xi_{i}^{t+1..t'}=\xi_{j}^{t+1..t'})\Rightarrow(a_{i}^{t+1..t'}=a_{j}^{t+1..t'}),~\forall t'\in[t+1,H],~\forall i,j\in[1,r]\label{eq:multistage_deterministic_equivalent_nonanticipativity}
\end{align}
The nonanticipativity constraints (\ref{eq:multistage_deterministic_equivalent_nonanticipativity}) state that, when 2 realizations $\xi_{i}^{t+1..H}$ and $\xi_{j}^{t+1..H}$ share a same prefix from $t+1$ to $t'$, the corresponding actions must be equal \cite{shapiro2009lectures}.

Solving eq. (\ref{eq:exact_two_stage}) is computationally intractable for two reasons.
First, since the number $r$ of possible realizations of $\boldsymbol\xi^{t+1..H}$ is exponential in the number of vertices and in the remaining horizon size $H-t$, considering every possible scenario is intractable in practice. We therefore consider a smaller set of $\alpha$ scenarios $S=\{{s}_{1},...,{s}_{\alpha}\}$ such that each scenario $s_i\in S$ is a realization of $\boldsymbol\xi^{t+1..H}$, {\em i.e.}, $\forall i\in [1,\alpha], s_i\in \boldsymbol\xi^{t+1..H}$. This set $S$ is obtained by Monte Carlo sampling \cite{asmussen2007stochastic}. All elements of $S$ share the same probability, {\em i.e.}, $p(\boldsymbol\xi^{t+1..H} = s_1) = \ldots = p(\boldsymbol\xi^{t+1..H} = s_\alpha)$.

Second, solving eq. (\ref{eq:exact_two_stage}) basically involves solving to optimality problem $\cal Q$ for each possible action $a^t\in A(\xi^t)$. Each problem $\cal Q$ involves solving a VRPTW for each possible scenario of $S$, while ensuring nonanticipativity constraints between the different solutions. As the VRPTW problem is an $\cal NP$-hard problem, we propose to compute an upper bound $\overline{\cal Q}$ of $\cal Q$ based on a given sequence $a_R^{t+1..H}$ of future route actions. Because we impose the sequence $a^{t+1..H}_R$, the set of possible actions at time $t$ is limited to those directly compatible with it, denoted $\tilde{A}(\xi^{t},a^{t+1..H}_R)\subseteq A(\xi^{t})$. That limitation enforces $\omega(a^{0..H})<+\infty$.  This finally leads to the GSA decision rule:
\begin{equation}
\hspace{-8em}\text{\emph{(GSA)}}\hspace{7em}
a^{t}=\argmin_{a^{t}\in \tilde{A}(\xi^{t},a^{t+1..H}_R)}  {\overline{\cal Q}}(a^{0..t},a^{t+1..H}_R,S)\label{eq:approx_two_stage}
\end{equation}
which, provided realization $\xi^{t}$, sampled scenarios $S$ and future route actions $a_R^{t+1..H}$, selects the action $a^t$ that minimizes the expected approximate cost over scenarios $S$. Notice that  almost all the anticipative efficiency of the GSA decision rule relies on the sequence $a_R^{t+1..H}$, which directly affects the quality of the upper bound $\overline{\cal Q}$.

\subsubsection{Sequence $a_R^{t+1..H}$ of future route actions.}
This sequence is used to compute an upper bound of $\cal Q$. For each time $t'\in [t+1,H]$, the route action $a_R^{t'}$ only contains operational decisions related to vehicle routing ({\em i.e.}, for each vehicle, travel towards a vertex, or wait at its current position) and does not contain decisions related to requests ({\em i.e.}, request acceptance or rejection). The more flexible $a_R^{t'}$ with respect to $S$, the better the bound $\overline{\cal Q}$. We describe in Section \ref{sec:Solving-DSVRPTW} how a flexible sequence is computed through local search.

\subsubsection{Computation of an upper bound $\overline{\cal Q}$ of $\cal Q$.}
Algorithm \ref{alg:The-L-approximation-function} depicts the  computation of an upper bound $\overline{\cal Q}$ of $\cal Q$ given a sequence $a_R^{t+1..H}$ of route actions consistent with past actions $a^{0..t}$.
\begin{algorithm}[b]
Precondition: $a_R^{t+1..H}$ is a sequence of route actions consistent with $a^{0..t}$\\
\For{each scenario $s_{i}\in S$}{
	${\it nbRejected}[i]\leftarrow0$;~ 
	$b^{0..t}\leftarrow a^{0..t}$;~ 
	$b^{t+1..H}\leftarrow a_R^{t+1..H}$\\

	\For{$t'\in [t+1..H]$}{
		\For{each request $(j,t')$ revealed at time $t'$ for a vertex $j$ in scenario $s_i$}{
			${c}^{t'..H}\leftarrow trytoServe((j,t'),b^{t'..H})$\label{line:mod} \\
			\textbf{if} ${b}^{t+1..t'-1}\cdot c^{t'..H}$ is feasible
			\textbf{then} ${b}^{t'..H}\leftarrow {c}^{t'..H}$\\
			\textbf{else} add the decision {\em reject(j,t')} to $b^{t'}$ and increment ${\it nbRejected}[i]$
		}
	}
}
\Return{$\frac{1}{|S|}\cdot \sum_{s_i\in S}{\it nbRejected}[i]$}
\protect\caption{The $\overline{\cal Q}(a^{0..t},a^{t+1..H}_R,S)$ approximation function\label{alg:The-L-approximation-function}}
\end{algorithm}
For each scenario $s_i$ of $S$, Algorithm \ref{alg:The-L-approximation-function} builds a sequence $b^{0..H}$ for $s_i$, which starts with $a^{0..t}$, and whose end $b^{t+1..H}$ is computed from $a_R^{t+1..H}$ in a greedy way. At each time $t'\in [t+1..H]$, each request revealed at time $t'$ in scenario $s_i$ is accepted if it is possible to modify $b^{t'..H}$ so that one vehicle can service it; it is rejected otherwise. One can consider $b^{t'..H}$ as being a set of vehicle routes, each defined by a sequence of planned vertices. Each planned vertex comes with specific decisions: a waiting time and whether a service is performed. In this context, $trytoServe$ performs a deterministic linear time modification of $b^{t'..H}$ such that $(j,t')$ corresponds to the insertion of the vertex $j$ in one of the routes defined by $b^{t'..H}$, at the best position with respect to VRPTW constraints and travel times, without modifying the order of the remaining vertices.
At the end, Algorithm \ref{alg:The-L-approximation-function} returns the average number of rejected requests for all scenarios. Note that, when modifying a sequence of actions so that a request can be accepted (line \ref{line:mod}), actions $b^{t'..H}$ can be modified, but $b^{0..t'-1}$ are not modified. This ensures that $\overline{{\cal Q}}$ preserves the nonanticipativity constraints. Indeed, the fact that two identical scenarios prefixes could be assigned two different subsequences of actions implies that either $trytoServe((j,t'),b^{t'..H})$ is able to modify an action $b^{t<t'}$ or is a nondeterministic function. In both cases, there is a contradiction. Finally, notice that contrary to other local search methods based on Monte Carlo simulation as in \cite{Ghiani2009,Schilde2011}, GSA considers the whole timing horizon when evaluating a first-stage solution against a scenario.

\subsubsection{Comparison to MSA}

GSA has two major differences with MSA. Given a set of scenarios, GSA maintains only one solution, namely the sequence $a^{t+1..H}_R$, that best suits to a pool of scenarios whilst MSA computes a set of solutions, each specialized to one scenario from the pool. Furthermore, by preserving nonanticipativity GSA approximates the multistage problem of equations (\ref{eq:multistage},\ref{eq:exact_two_stage}). In contrary, MSA relaxes these constraints and therefore approximates the two-stage problem (\ref{eq:twostage}) \cite{VanHentenryck2009}. 

In particular, given a pool of scenarios obtained by Monte Carlo sampling, MSA Expectation Algorithm \ref{alg:expectation-algo} reformulates eq.  (\ref{eq:twostage}) as a \emph{sample average approximation} (SAA, \cite{ahmed2002sample,verweij2003sample}) problem. The SAA tackles each scenario as a separate deterministic problem. For a specific scenario $\xi^{t+1..H}$, it considers the recourse cost of a solution starting with actions $a^{0..t}$. Because the scenarios are not linked by nonanticipativity constraints, two scenarios $i$ and $j$ that share the same prefix $\xi^{t+1..t'}$ can actually be assigned two solutions performing completely different actions $a_{i}^{0..t'}$ and $a_{j}^{0..t'}$, for some $t'>t$. The evaluation of action $a^{t}$ over the set of scenarios is therefore too optimistic, leading to a suboptimal choice. By definition, the \emph{Regret} algorithm approximates the Expectation algorithm. The \emph{Regret} algorithm then also approximates a two-stage problem. The \emph{Consensus} algorithm selects the most suggested action among plans of the pool. By selecting the most frequent action in the pool, \emph{Consensus} somehow encourages nonanticipation. However, the nonanticipativity constraints are not enforced as each scenario is solved separately. \emph{Consensus} also approximates a two-stage problem.

\section{Solving the Dynamic and Stochastic VRPTW\label{sec:Solving-DSVRPTW}}
GSA alone does not permit to solve a DS-VRPTW instance. In this section, we now show how the decision rule, as defined in eq. \ref{eq:approx_two_stage}, can be embedded in an online algorithm that solves the DS-VRPTW. Finally, we present the different waiting and relocation strategies we exploit, including a new waiting strategy.

\subsection{Embedding GSA\label{sub:Embedding-GSA}}

In order to solve the DS-VRPTW, we design Algorithm \ref{alg:LS-GSA}, which embeds the GSA decision rule. 

\begin{algorithm}[t!]
Initialize $S$ with $\alpha$ scenarios and compute initial solution $a_R^{1..H}$ w.r.t. known requests\\
$t \leftarrow 1$; \\
\While{real time has not reached the end of the time horizon}{
	\tcc{Beginning of the time unit}
	$(a^t,a_R^{t+1..H})\leftarrow${\tt handleRequests}$(a^{0..t-1},a_R^{t..H},\xi^t)$\\
	execute action $a^t$ and update the pool $S$ of scenarios w.r.t. to $\xi^t$\\
	
	\tcc{Remaining of the time unit}
	\While{real time has not reached the end of time unit $t$}{
		$b_R^{t+1..H}\leftarrow{\tt shakeSolution}(a_R^{t+1..H})$\\
		\lIf{$\overline{\cal Q}(a^{0..t},b_R^{t+1..H},S)<\overline{\cal Q}(a^{0..t},a_R^{t+1..H},S)$}{
			$a_R^{t+1..H}\leftarrow b_R^{t+1..H}$
		}
		\If{the number of iterations since the last re-initialization of $S$ is equal to $\beta$}{
			Re-initialize the pool $S$ of scenarios w.r.t. $\boldsymbol\xi^{t+1..H}$
		}
	}	
	$t\leftarrow t+1$~~~~ \tcc{Skip to next time unit}
	
}
\textbf{Function} \SetKwFunction{proc}{handleRequests}
\proc{$a^{0..t-1},a_R^{t..H},\xi^t$} \\
{
	$b^{0..t-1}\leftarrow a^{0..t-1}$;~
	$b^{t..H}\leftarrow a_R^{t..H}$\\
	\For{each request revealed for a vertex $j$ in realization $\xi^t$}{
		\eIf{we find, in less than $\delta_{ins}$, how to modify $b^{t..H}$ s.t. request $(j,t)$ is served}
		{modify $b^{t..H}$ to accept request $(j,t)$}
		{modify $b^{t..H}$ to reject request $(j,t)$}
	}
\Return{$(b^t,b^{t+1..H})$}
}
\protect\caption{LS-based GSA\label{alg:LS-GSA}}
\end{algorithm}

\paragraph{Main Algorithm.}
It is parameterized by: $\alpha$ which determines the size of the pool $S$ of scenarios; $\beta$ which determines the frequency for re-initializing $S$; and $\delta_{ins}$ which limits the time spent for trying to insert a request in a sequence. 

It runs in \emph{real time}. It is started before the beginning of the time horizon, in order to compute an initial pool $S$ of $\alpha$ scenarios and an initial solution $a_R^{1..H}$ with respect to offline requests (revealed at time 0). It runs during the whole time horizon, and loops on lines 3 to 11. It is stopped when reaching the end of the time horizon. The \emph{real time} is discretized in $H$ time units, and the variable $t$ represents the current time unit: It is incremented when real time exceeds the end of the $t^{th}$ time unit. In order to be correct, Algorithm \ref{alg:LS-GSA} requires the real computation time of lines 4 to 11 to be smaller than the real time spent in one time unit. This is achieved by choosing suitable values for parameters $\alpha$ and $\delta_{ins}$.

Lines 4 and 5 describe what happens whenever the algorithm enters a new time unit: Function {\tt handleRequests} (described below) chooses the next action $a^t$ and updates $a_R^{t+1..H}$; Finally, $S$ is updated such that it stays coherent with respect to realization $\xi^t$. Each scenario $\xi^{t..H}\in S$ is composed of a sequence of sampled requests. To each customer region $i$ is associated an upper bound $\overline{r}_i=\min ( l_0-t_{i,0}-d_i,l_i-t_{0,i} )$ on the time unit at which a request can be revealed in that region, like in \cite{bent2004scenario}. That constraint prevents tricky or inserviceable requests to be sampled. At time $t$, a sampled request $(i,t)$ which doesn't appear in $\xi^t$ is either removed if $t \ge \overline{r}_i$ or randomly delayed in $\xi^{t+1..H}\in S$ otherwise.

The algorithm spends the rest of the time unit to iterate over lines 7 to 10, in order to improve the sequence of future route actions $a_R^{t+1..H}$. We consider a hill climbing strategy: The current solution $a_R^{t+1..H}$ is shaked to obtain a new candidate solution $b_R^{t+1..H}$, and if this solution leads to a better upper bound $\overline{\cal Q}$ of ${\cal Q}$, then it becomes the new current solution. Shaking is performed by the {\tt shakeSolution} function. This function considers different neighborhoods, corresponding to the following move operators: relocate, swap, inverted 2-opt, and cross-exchange (see \cite{kindervater1997vehicle,Taillard1997} for complete descriptions). 
As explained in Section \ref{sub:Waiting-and-Relocation}, depending on the chosen waiting and relocation strategy, additional move operators are exploited. At each call to the {\tt shakeSolution} function, the considered move operator is changed, such that the operators are equally selected one after another in the list. Every $\beta$ iterations, the pool $S$ of scenarios is re-sampled (lines 9-10). This re-sampling introduces diversification as the upper bound computed by $\overline{\cal Q}$ changes. We therefore do not need any other meta-heuristic such as Simulated Annealing.

\paragraph{Function {\tt handleRequest}} is called at the beginning of a new time unit $t$, to compute action $a^t$ in light of online requests (if any). It implements the GSA decision rule defined in eq. (\ref{eq:approx_two_stage}). The function considers each request revealed at time $t$ for a vertex $j$, in a sequential way. For each request, it tries to insert it into the sequence $a_R^{t..H}$ ({\em i.e.}, modify the routes so that a vehicle visits $j$ during its time window). As in {\tt shakeSolution}, local search operations are performed during that computation. The time spent to find a feasible solution including the new request is limited to $\delta_{ins}$. If such a feasible solution is found, then the request is accepted, otherwise it is rejected. If there are several online requests for the same discretized time $t$, we process these requests in their real-time order of arrival, and we assume that all requests are revealed at different real times.


\subsection{Waiting and Relocation strategies\label{sub:Waiting-and-Relocation}}

As defined in section 2, a vehicle that just visited a vertex usually has the choice between traveling right away to the next planned vertex or first waiting for some time at its current position. Unlike in the static (and deterministic) case, in the dynamic (and stochastic) VRPTW these choices may have a significant impact on the solution quality. 

Waiting and relocation strategies have attracted a great interest on dynamic and stochastic VRP's. In this section, we present and describe how waiting and relocation strategies are integrated to our framework, including a new waiting strategy called \emph{relocation-only}.

\subsubsection{Relocation strategies}
Studies in \cite{Bertsimas1991,Bertsimas1993} already showed that for
a dynamic VRP with no stochastic information, it is optimal to relocate the vehicle(s) either to the center (in case of single-vehicle) or to strategical points (multiple-vehicle case) of the service region.
The idea evolved and has been successfully adapted to routing problems
with customer stochastic information, in reoptimization approaches
as well as sampling approaches.

Relocation strategies explore solutions obtained when allowing a vehicle
to move towards a customer vertex even if there is no request received
for that vertex at the current time slice. Doing so, one recognizes
the fact that, in the context of dynamic and stochastic vehicle routing,
a higher level of anticipation can be obtained by considering to reposition
the vehicle after having serviced a request to a more stochastically
fruitful location. Such a relocation strategy has already been applied
to the DS-VRPTW in \cite{Bent2007}.

\subsubsection{Waiting strategies}

In a dynamic context, the planning of a vehicle usually contains more
time than needed for traveling and servicing requests. When it finishes
to service a request, a vehicle has the choice between waiting for
some time at its location or leaving for the next planned vertex.
A good strategy for deciding where and how long to wait can potentially
help at anticipating future requests and hence increase 
the dynamic performances. We consider three existing waiting
strategies and introduce a new one:
\begin{itemize}
\item \emph{Drive-First ($DF$)}: The basic strategy aims at leaving each serviced
request as soon as possible, and possibly wait at the next vertex
before servicing it if the vehicle arrives before its time window.

\item \emph{Wait-First ($WF$)}: Another classical waiting strategy consists in delaying as much as possible the service time of every planned requests, without violating their time windows. After having serviced a request, the vehicle hence waits as long as possible before moving to the next planned request.
\item \emph{Custom-Wait ($CW$)}: A more tailored waiting strategy aims at controlling the waiting time at each vertex, which becomes part of the online decisions.
\item \emph{Relocation-Only waiting ($RO$)}: In order to take maximum benefit of relocation strategy while avoiding the computational overhead due to additional decision variables involved in custom waiting, we introduce a new waiting strategy.
It basically applies \emph{drive-first} scheduling to every request and
then applies \emph{wait-first} waiting only to those requests that follow
a relocation one. By doing so, a vehicle will try to arrive as soon
as possible at a planned \emph{relocation request}, and wait there
as long as possible. In contrary, it will spend as less time as possible
at non-relocation request vertices. Note that if it is not coupled to a relocation strategy, $RO$  reduces to $DF$. Furthermore, $RO$ also reduces to the dynamic waiting strategy described in \cite{Mitrovic-Minic2004a} if we define the service zones as being delimited by relocation requests.
However, our strategy differs by the fact that service zones in our
approach are computed in light of stochastic information instead of
geometrical considerations.
\end{itemize}
Depending on the waiting strategy we apply and whether we use relocation
or not, additional LS move operators are exploited. Specifically,
among the waiting strategies, only \emph{custom-wait} requires additional
move operators aiming at either increasing or decreasing the waiting time at a random planned vertex. 
\emph{Relocation} also requires two additional move operators that modify a given solution by either inserting or removing a relocation action at a random vertex.

\section{Experimentations\label{sec:Experimentations}}

We now describe our experimentations and compare our results with
those of the state of the art MSA algorithm of \cite{bent2004scenario}.

\subsection{Algorithms}\label{sec:algos}

Different versions of Algorithm \ref{alg:LS-GSA} have been experimentally assessed, depending on which waiting strategy is implemented and whether in addition we use the relocation strategy or not.

Surprisingly, the \emph{wait-first} waiting strategy, as well as its version including \emph{relocation}, produced very bad results in comparison to other strategies, rejecting more than twice more online requests in average. Because of its computational overhead, the \emph{custom-wait} strategy also produced bad results, even with relocation. For conciseness we therefore do not report these strategies in the result plots.

The 3 different versions of Algorithm \ref{alg:LS-GSA} we thus consider are the following: GSA\emph{df}, which stands for GSA with \emph{drive-first} waiting strategy, GSA\emph{dfr} which stands for GSA with \emph{drive-first} and \emph{relocation} strategies, and finally GSA\emph{ro} with means GSA using \emph{relocation-only} strategy. Recall that, by definition, the \emph{relocation-only} strategy involves relocation.
In addition to those 3 algorithms, as a baseline we consider the GLS\emph{df} algorithm, which stands for \emph{greedy local search} with \emph{drive-first} waiting. This algorithm is similar to the dynamic LS described in \cite{Schilde2011}, to which we coupled a Simulated Annealing metaheuristic. In this algorithm, stochastic information about future request is not taken into account and a neighboring solution is solely evaluated by its total travel cost. 

Finally, GSA and GLS are compared to two MSA algorithms,
namely MSA\emph{d} and MSA\emph{c} depending on whether the \emph{travel distance} or
the \emph{consensus function} are used as ranking functions.

\subsection{Benchmarks}

The selected benchmarks are borrowed from \cite{bent2004scenario} which considers a set
of benchmarks initially designed for the static and deterministic
VRPTW in \cite{Solomon1987}, each of these containing 100
customers. In our stochastic and dynamic context, each customer becomes
a request region, where dynamic requests can occur during the online
execution. 

The original problems from \cite{bent2004scenario} are divided into
4 classes of 15 instances. Each class is characterized by its degree
of dynamism (DOD, the ratio of the number of dynamic requests revealed at time $t>0$ over
the number of a priori request known at time $t=0$) and whether the dynamic requests are
known early or lately along the online execution. The time horizon $H=480$ is divided into 3 time slices. A request is said to be early if it is revealed during the first time slice $t\in[1,160]$. A late request is revealed during the second time slice $t\in[161,320]$. There is no request revealed during the third time slice $t\in[321,480]$, but the vehicles can use it to perform customer operations.

In Class 1 there are many initial requests,
many early requests and very few late requests. Class 2 instances have many initial requests, very few
early requests and some late requests. Class 3 is a mix of classes 1 and 2. In Class 4, there are few initial requests, few early requests and many late requests. Finally, classes 1, 2 and 3 have an average DOD of
44\%, whilst Class 4 has an average DOD of 57\%.

In  \cite{Bent2007}, a fifth class is proposed with a higher DOD of 81\% in average.
Unfortunately, we were not able to get those Class 5 instances. We complete these classes by providing a sixth class of instance, with DOD of 100\%. Each instance hence contains no initial request, an early request with probability $0.3$ and a late request with probability $0.7$. 

Figure \ref{fig:instances} summarizes the different instance classes.

\begin{figure}
\centering
\begin{tabular}{|r|c|c|c|c|c|c|c|c|}
\hline
 & ~~DOD~~ & $t=0$ & ~~$t\in[1,160]$~~ & ~~$t\in[161,320]$~~ & ~~$t\in[321,480]$~~ \\\hline
Class 1,2,3 & 44\% & $P^0[i]=0.5$ & $P^{[1,160]}[i]=0.25$ & $P^{[161,320]}[i]=0.25$ & $P^{[321,480]}[i]=0$ \\\hline
Class 4 & 57\% & $P^0[i]=0.2$ & $P^{[1,160]}[i]=0.2$ & $P^{[161,320]}[i]=0.6$ & $P^{[321,480]}[i]=0$ \\\hline
Class 5 & 81\% & $P^0[i]=0.1$ & $P^{[1,160]}[i]=0.1$ & $P^{[161,320]}[i]=0.8$ & $P^{[321,480]}[i]=0$ \\\hline
Class 6 & 100\% & $P^0[i]=0$ & $P^{[1,160]}[i]=0.3$ & $P^{[161,320]}[i]=0.7$ & $P^{[321,480]}[i]=0$ \\\hline
\end{tabular}
\protect\caption{Summary of the test instances, grouped per degree of dynamism. $P^{[t,t']}[i]$ represents the probability that a request gets revealed during the time slice defined by interval $[t,t'].$ \label{fig:instances}}
\end{figure}

\subsection{Results}

Computations are performed on a cluster composed of 32 64-bits AMD
Opteron(tm) Processor 6284 SE cores, with CPU frequencies ranging
from 1400 to 2600 MHz. Executables were developed with C++ and compiled on a Linux Red Hat environment with GCC 4.4.7. Average results over 10 runs are reported. In  \cite{bent2004scenario}, 25 minutes of offline computation are allocated to MSA, in order to decide the first online action at time $t=1$. During online execution, each time unit within the time horizon was
executed during 7.5 seconds by the simulation framework. In order to compensate the technology difference, we decided in this study to allow only 10 minutes of offline computation and 4 seconds of online computation per time unit. Thereafter, in order
to highlight the contribution of the offline computation in our approach,
the amount of time allowed at pre-computation is increased to 60 minutes, 
while each time unit still lasts 4 seconds. According to preliminary experiments, 
both the size of the scenario pool and the resampling rate are set 
to $\alpha=\beta=150$ for all our algorithms except GLS\emph{df}. 

\begin{figure}
\centering
\includegraphics[width=0.9\columnwidth]{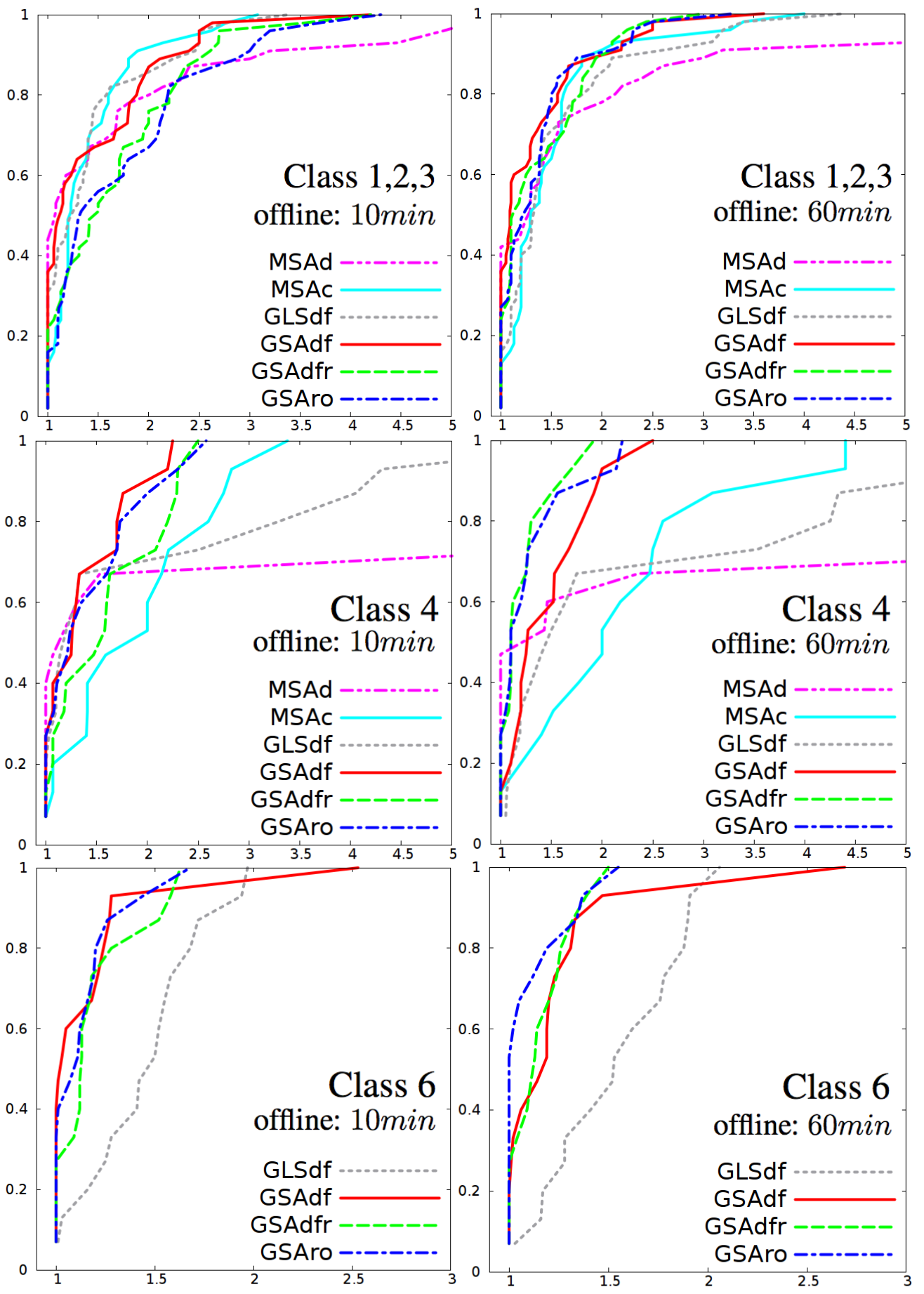}
\protect\caption{Performance profiles on classes [1, 2 ,3], Class 4 and Class 6 problem instances\label{fig:Profiles_All}}
\end{figure}

Figure \ref{fig:Profiles_All} gives a graphical representation of our algorithms results, through performance profiles. Performance profiles provide, for each algorithm, a cumulative distribution of its performance compared to other algorithms. For a given algorithm, a point $(x,y)$ on its curve means that, in $(100\cdot y)\%$ of the instances, this algorithm performed at most $x$ times worse than the best algorithm on each instance taken separately. Instances are grouped by DOD and by offline computation time. Classes 1, 2 and 3 have a DOD of 44\%, hence they are grouped together. 
An algorithm is strictly better than another one if its curve stays above the other algorithm's curve. For example on the 60min plot of Class 6, GLS\emph{df} is the worst algorithm in $95\%$ of Class 6 instances, outperforming GSA\emph{df} in the remaining $5\%$ (but not the other algorithms). On the other hand, provided these 60 minutes of offline computation, GSA\emph{ro} obtains the best results in $55\%$ of the instances, whereas only $30\%$ for GSA\emph{df} and GSA\emph{dfr}.
See \cite{dolan2002benchmarking} for a complete description of performance profiles. Detailed results are provided in the appendix.

Our algorithms compare fairly with MSA, especially on lately dynamic instances of Class 4. Given more offline computation, our algorithms get stronger, although that MSA benefits of the same offline time in every plots. Surprisingly, GLS\emph{df} performs well compared to other algorithms on classes 1,2 and 3. The low DOD that characterizes these instances tends to lower the contribution of stochastic knowledge against the computational power of GLS\emph{df}. Indeed, approximating the stochastic evaluation function over 150 scenarios is about $10^3$ times more expensive than GLS\emph{df} evaluation function. However, as the offline computation time and the DOD increase, stochastic algorithms tend to outperform their deterministic counterpart. 

We notice that the relocation strategy gets stronger as the offline computation time increases. This is due to the computational overhead induced by relocation vertices. GSA\emph{df} is then the good choice under limited offline computation time.  However, both GSA\emph{ro} and GSA\emph{dfr} tend to outperform the other strategies when provided enough offline computation and high DOD. 

As it contains no deterministic request, in Class 6 the offline computation is not applicable to those
algorithms that does not exploit the relocation strategy, i.e. GLS\emph{df} and GSA\emph{df}. Class 6 shows that, despite the huge difference in the number of iterations performed by GLS\emph{df} on one hand and stochastic algorithms on the other, the laters clearly outperform GLS\emph{df} under fully dynamic instances. We also notice in this highly dynamic context that GSA\emph{ro} tends to outperform GSA\emph{dfr} as offline computation increases, highlighting the anticipative contribution provided by the \emph{relocation-only} strategy, centering waiting times on relocation vertices.

\section{Conclusions\label{sec:Conclusions}}

We proposed GSA, a decision rule for dynamic and stochastic vehicle routing with time windows (DS-VRPTW), based on a stochastic programming heuristic approach. Existing related studies, such as MSA, simplify the problem as a two-stage problem by using sample average approximation. In contrary, the theoretical singularity of our method is to approximate a multistage stochastic problem through Monte Carlo sampling, using a heuristic evaluation function that preserves the nonanticipativity constraints. By maintaining one unique anticipative solution designed to be as flexible as possible according to a set of scenarios, our method differs in practice from MSA which computes as many solutions as scenarios, each being specialized for its associated scenario. Experimental results show that GSA produces competitive results with respect to state-of-the-art. This paper also proposes a new waiting strategy, \emph{relocation-only}, aiming at taking full benefit of relocation strategy. 

In a future study we plan to address a limitation of our solving algorithm which embeds GSA, namely the computational cost of its evaluation function. One possible direction would be to take more benefit of each evaluation, by spending much more computational effort in constructing neighboring solutions, e.g. by using Large Neighborhood Search \cite{shaw1998using}. Minimizing the operational cost, such as the total travel distance, is usually also important in stochastic VRPs. Studying the aftereffect when incorporating it as a second objective should be of worth. It is also necessary to consider other types of DS-VRPTW instances, such as problem sets closer to public or good transportation. Finally, the conclusions we made in section 2 about the shortcoming of a two-stage formulation (showed in Fig. 1) are theoretical only, and should be experimentally assessed. 

\section*{Acknowledgement}
Christine Solnon is supported by the LABEX IMU (ANR-10-LABX-0088) of Université de Lyon, within the program "Investissements d'Avenir" (ANR-11-IDEX-0007) operated by the French National Research Agency (ANR). This research is also partially supported by the UCLouvain Action de Recherche Concertée ICTM22C1.


\section*{Appendix: Detailed experimental results}
The present appendix provides detailed experimental results on the different instance classes. 

Tables \ref{tab:results_class1}, \ref{tab:results_class2}, \ref{tab:results_class3} and \ref{tab:results_class4} show detailed results on each algorithms for classes 1-4, with both 10 and 60 minutes allowed at offline computation. Note that both MSA\emph{d} and MSA\emph{c} should only be compared with our algorithms when allowing 10 minutes at offline computation. Table \ref{tab:results_class6} shows the results obtained on Class 6 problem instances. Since the instances belonging to Class 6 contain no deterministic request, offline computation can only be performed on these instance when allowing relocation.

\begin{table}[p]
\centering
\begin{tabular}{ l  r  c c    r	 c   r   c c c   r   c  r   c c c   }
Class 1 &  &   \multicolumn{8}{c}{10 min. offline computation} &  &  \multicolumn{5}{c}{60 min. offline comput.} \\
         	&   & \multicolumn{2}{c}{MSA}  &     & \multicolumn{1}{c}{GLS}        & 			& \multicolumn{3}{c}{GSA}                  &                   & \multicolumn{1}{c}{GLS}   &      & \multicolumn{3}{c}{GSA}                                                                              \\
  Instance    & \hspace{3em}  & \hspace{0.2em} \textit{d\phantom{fr}} \hspace{0.2em} & \hspace{0.2em} \textit{c\phantom{f}} \hspace{0.2em}  & \hspace{0.7em}  & \hspace{0.2em} \textit{df} \hspace{0.2em}  & \hspace{0.7em}  & \hspace{0.2em} \textit{df} \hspace{0.2em}  & \hspace{0.2em} \textit{dfr} \hspace{0.2em}  & \hspace{0.2em} \textit{ro} \hspace{0.2em} & \hspace{2.5em}  & \hspace{0.2em} \textit{df} \hspace{0.2em} & \hspace{0.7em}  & \hspace{0.2em} \textit{df} \hspace{0.2em}  & \hspace{0.2em} \textit{dfr} \hspace{0.2em}  & \hspace{0.2em} \textit{ro} \hspace{0.2em}  \\
\hline
RC 101-1 &  & 1.0                            & 0.6                            &  & \textbf{0.5}                    &  & 2.6                             & 1.1                              & 3.3                             &  & 1.2                             &  & \textbf{1.1}                    & 2.9                              & 1.2                             \\
RC 101-2 &  & 1.8                            & 2.6                            &  & \textbf{1.7}                    &  & 2.1                             & 2.8                              & 2.6                             &  & \textbf{1.3}                    &  & 1.6                             & 1.4                              & 1.6                             \\
RC 101-3 &  & 1.8                            & \textbf{1.0}                   &  & 1.7                             &  & 2.8                             & 3.0                              & 3.3                             &  & 1.8                             &  & 2.0                             & 2.3                              & \textbf{1.6}                    \\
RC 101-4 &  & \textbf{0.0}                   & 0.2                            &  & 0.5                             &  & 0.8                             & 1.7                              & 2.7                             &  & \textbf{0.4}                    &  & 1.5                             & 1.1                              & 1.1                             \\
RC 101-5 &  & \textbf{0.4}                   & 1.0                            &  & 0.7                             &  & 2.5                             & 2.5                              & 2.7                             &  & \textbf{0.8}                    &  & 0.9                             & 1.5                              & 2.2                             \\
\textit{Avg}      &  & \textbf{1.0}                   & 1.1                            &  & \textbf{1.0}                    &  & 2.2                             & 2.2                              & 2.9                             &  & \textbf{1.1}                    &  & 1.4                             & 1.8                              & 1.5                             \\
\hline
RC 102-1 &  & 2.0                            & 2.4                            &  & \textbf{0.8}                    &  & 2.0                             & 2.6                              & 2.9                             &  & \textbf{1.1}                    &  & 2.5                             & 2.0                              & 1.5                             \\
RC 102-2 &  & \textbf{0.4}                   & 0.8                            &  & 1.0                             &  & 0.8                             & 1.4                              & 1.9                             &  & 1.9                             &  & \textbf{0.8}                    & 1.4                              & 0.8                             \\
RC 102-3 &  & 1.0                            & 0.8                            &  & 0.8                             &  & \textbf{0.7}                    & 1.7                              & 1.2                             &  & 0.5                             &  & \textbf{0.3}                    & 0.4                              & 0.8                             \\
RC 102-4 &  & 1.2                            & 1.4                            &  & 1.6                             &  & \textbf{0.5}                    & 0.4                              & 0.8                             &  & 1.7                             &  & 0.5                             & \textbf{0.4}                     & 0.5                             \\
RC 102-5 &  & 1.2                            & 0.6                            &  & \textbf{0.4}                    &  & 0.6                             & 0.3                              & 0.5                             &  & 0.6                             &  & 0.4                             & \textbf{0.0}                     & 0.1                             \\
\textit{Avg}      &  & 1.2                            & 1.2                            &  & \textbf{0.9}                    &  & \textbf{0.9}                    & 1.3                              & 1.5                             &  & 1.2                             &  & 0.9                             & 0.8                              & \textbf{0.7}                    \\
\hline
RC 104-1 &  & \textbf{0.0}                   & 0.2                            &  & 0.1                             &  & \textbf{0.0}                    & 0.7                              & 1.1                             &  & \textbf{0.1}                    &  & \textbf{0.1}                    & \textbf{0.1}                     & 0.4                             \\
RC 104-2 &  & \textbf{0.0}                   & \textbf{0.0}                   &  & \textbf{0.0}                    &  & \textbf{0.0}                    & \textbf{0.0}                     & 0.3                             &  & 0.2                             &  & \textbf{0.0}                    & 0.1                              & \textbf{0.0}                    \\
RC 104-3 &  & \textbf{0.0}                   & \textbf{0.0}                   &  & \textbf{0.0}                    &  & \textbf{0.0}                    & \textbf{0.0}                     & \textbf{0.0}                    &  & \textbf{0.0}                    &  & \textbf{0.0}                    & \textbf{0.0}                     & \textbf{0.0}                    \\
RC 104-4 &  & \textbf{0.0}                   & 0.2                            &  & \textbf{0.0}                    &  & \textbf{0.0}                    & 0.1                              & 0.1                             &  & \textbf{0.1}                    &  & \textbf{0.1}                    & \textbf{0.1}                     & 0.2                             \\
RC 104-5 &  & \textbf{0.0}                   & \textbf{0.0}                   &  & \textbf{0.0}                    &  & \textbf{0.0}                    & \textbf{0.0}                     & \textbf{0.0}                    &  & \textbf{0.0}                    &  & \textbf{0.0}                    & \textbf{0.0}                     & \textbf{0.0}                    \\
\textit{Avg}      &  & \textbf{0.0}                   & \textbf{0.0}                   &  & \textbf{0.0}                    &  & \textbf{0.0}                    & 0.2                              & 0.3                             &  & 0.1                             &  & \textbf{0.0}                    & 0.1                              & 0.1                             \\
\hline
AVG      &  & 0.7                            & 0.8                            &  & \textbf{0.6}                    &  & 1.0                             & 1.2                              & 1.6                             &  & \textbf{0.8}                    &  & \textbf{0.8}                    & 0.9                              & \textbf{0.8}                    \\
                                                
\end{tabular}
\protect\caption{ Detailed results for Class 1 problem instances. Averages over 10 runs.\label{tab:results_class1}}
\end{table}

\begin{table}[p]
\centering
\begin{tabular}{ l  r  c c    r	 c   r   c c c   r   c  r   c c c   }
Class 2 &  &   \multicolumn{8}{c}{10 min. offline computation} &  &  \multicolumn{5}{c}{60 min. offline comput.} \\
         	&   & \multicolumn{2}{c}{MSA}  &     & \multicolumn{1}{c}{GLS}        & 			& \multicolumn{3}{c}{GSA}                  &                   & \multicolumn{1}{c}{GLS}   &      & \multicolumn{3}{c}{GSA}                                                                              \\
  Instance    & \hspace{3em}  & \hspace{0.2em} \textit{d\phantom{fr}} \hspace{0.2em} & \hspace{0.2em} \textit{c\phantom{f}} \hspace{0.2em}  & \hspace{0.7em}  & \hspace{0.2em} \textit{df} \hspace{0.2em}  & \hspace{0.7em}  & \hspace{0.2em} \textit{df} \hspace{0.2em}  & \hspace{0.2em} \textit{dfr} \hspace{0.2em}  & \hspace{0.2em} \textit{ro} \hspace{0.2em} & \hspace{2.5em}  & \hspace{0.2em} \textit{df} \hspace{0.2em} & \hspace{0.7em}  & \hspace{0.2em} \textit{df} \hspace{0.2em}  & \hspace{0.2em} \textit{dfr} \hspace{0.2em}  & \hspace{0.2em} \textit{ro} \hspace{0.2em}  \\
\hline

RC 101-1 &  & \textbf{0.0} & 0.2          &  & 0.4          &  & 1.5          & 1.2          & 2.2          &  & \textbf{0.1} &  & 0.6          & 0.9          & 1.5          \\
RC 101-2 &  & \textbf{1.0} & 1.4          &  & 1.9          &  & 2.3          & 4.2          & 3.2          &  & \textbf{1.7} &  & 2.3          & 3.2          & 2.1          \\
RC 101-3 &  & \textbf{0.0} & \textbf{0.0} &  & 1.5          &  & 3.2          & 3.2          & 3.3          &  & 2.2          &  & 2.6          & \textbf{2.0} & 2.3          \\
RC 101-4 &  & \textbf{0.6} & 0.8          &  & 1.2          &  & 3.2          & 4.7          & 3.9          &  & \textbf{1.9} &  & 3.0          & 2.6          & 2.7          \\
RC 101-5 &  & 1.6          & \textbf{1.4} &  & 1.6          &  & 2.5          & 3.2          & 1.7          &  & \textbf{1.2} &  & 1.4          & 1.7          & 2.1          \\
\textit{Avg}      &  & \textbf{0.6} & 0.8          &  & 1.3          &  & 2.5          & 3.3          & 2.9          &  & \textbf{1.4} &  & 2.0          & 2.1          & 2.1          \\
\hline
RC 102-1 &  & \textbf{0.0} & 0.4          &  & \textbf{0.0} &  & 1.1          & 1.2          & 1.0          &  & \textbf{0.0} &  & 0.9          & 0.7          & 0.4          \\
RC 102-2 &  & \textbf{0.6} & 1.2          &  & \textbf{0.6} &  & \textbf{0.6} & 1.1          & 1.0          &  & \textbf{0.7} &  & 1.0          & 0.8          & 0.8          \\
RC 102-3 &  & 2.0          & 2.0          &  & \textbf{0.4} &  & 1.5          & 1.1          & 1.3          &  & \textbf{1.0} &  & 1.1          & 1.2          & \textbf{1.0} \\
RC 102-4 &  & \textbf{0.2} & 0.4          &  & 0.9          &  & 0.4          & 0.4          & 1.1          &  & 0.6          &  & \textbf{0.2} & 0.5          & 0.8          \\
RC 102-5 &  & 2.6          & 2.8          &  & \textbf{2.1} &  & 2.3          & 2.3          & 2.7          &  & 1.5          &  & 1.5          & 1.5          & \textbf{1.3} \\
\textit{Avg}      &  & 1.1          & 1.4          &  & \textbf{0.8} &  & 1.2          & 1.2          & 1.4          &  & \textbf{0.8} &  & 0.9          & 0.9          & 0.9          \\
\hline
RC 104-1 &  & 6.2          & 3.0          &  & 2.2          &  & 0.5          & 0.6          & \textbf{0.3} &  & 2.4          &  & \textbf{0.0} & 0.1          & 0.1          \\
RC 104-2 &  & 5.4          & 2.6          &  & 2.6          &  & 0.3          & 2.5          & 2.9          &  & 2.4          &  & \textbf{0.2} & \textbf{0.2} & 0.6          \\
RC 104-3 &  & 2.0          & 0.8          &  & 1.1          &  & \textbf{0.0} & \textbf{0.0} & 0.8          &  & 0.9          &  & \textbf{0.0} & \textbf{0.0} & \textbf{0.0} \\
RC 104-4 &  & 0.8          & 0.6          &  & 0.1          &  & \textbf{0.0} & \textbf{0.0} & \textbf{0.0} &  & 0.4          &  & \textbf{0.0} & \textbf{0.0} & \textbf{0.0} \\
RC 104-5 &  & 4.2          & 0.2          &  & 1.5          &  & \textbf{0.1} & 0.2          & \textbf{0.1} &  & 0.7          &  & \textbf{0.0} & \textbf{0.0} & 0.1          \\
\textit{Avg}      &  & 3.7          & 1.4          &  & 1.5          &  & \textbf{0.2} & 0.7          & 0.8          &  & 1.4          &  & \textbf{0.0} & 0.1          & 0.2          \\
\hline
AVG      &  & 1.8          & \textbf{1.2} &  & \textbf{1.2} &  & 1.3          & 1.7          & 1.7          &  & 1.2          &  & \textbf{1.0} & \textbf{1.0} & \textbf{1.0}

\end{tabular}
\protect\caption{ Detailed results for Class 2 problem instances. Averages over 10 runs. \label{tab:results_class2}}
\end{table}

\begin{table}[p]
\centering
\begin{tabular}{ l  r  c c    r	 c   r   c c c   r   c  r   c c c   }
Class 3 &  &   \multicolumn{8}{c}{10 min. offline computation} &  &  \multicolumn{5}{c}{60 min. offline comput.} \\
         	&   & \multicolumn{2}{c}{MSA}  &     & \multicolumn{1}{c}{GLS}        & 			& \multicolumn{3}{c}{GSA}                  &                   & \multicolumn{1}{c}{GLS}   &      & \multicolumn{3}{c}{GSA}                                                                              \\
  Instance    & \hspace{3em}  & \hspace{0.2em} \textit{d\phantom{fr}} \hspace{0.2em} & \hspace{0.2em} \textit{c\phantom{f}} \hspace{0.2em}  & \hspace{0.7em}  & \hspace{0.2em} \textit{df} \hspace{0.2em}  & \hspace{0.7em}  & \hspace{0.2em} \textit{df} \hspace{0.2em}  & \hspace{0.2em} \textit{dfr} \hspace{0.2em}  & \hspace{0.2em} \textit{ro} \hspace{0.2em} & \hspace{2.5em}  & \hspace{0.2em} \textit{df} \hspace{0.2em} & \hspace{0.7em}  & \hspace{0.2em} \textit{df} \hspace{0.2em}  & \hspace{0.2em} \textit{dfr} \hspace{0.2em}  & \hspace{0.2em} \textit{ro} \hspace{0.2em}  \\
\hline

RC 101-1     &  & \textbf{0.6} & 0.8 &  & 0.9          &  & 1.9          & 2.8          & 2.5          &  & \textbf{0.9} &  & 1.5          & 2.1          & 1.8          \\
RC 101-2     &  & 2.2          & 1.4 &  & 0.9          &  & 1.4          & 2.7          & \textbf{1.2} &  & 1.0          &  & \textbf{0.5} & 1.2          & 1.2          \\
RC 101-3     &  & \textbf{0.6} & 0.8 &  & 1.0          &  & 2.0          & 2.6          & 2.7          &  & \textbf{1.1} &  & 2.5          & 1.6          & 1.5          \\
RC 101-4     &  & \textbf{0.6} & 1.0 &  & 1.3          &  & 2.1          & 2.1          & 1.8          &  & \textbf{0.9} &  & 1.5          & 1.9          & 1.4          \\
RC 101-5     &  & \textbf{0.0} & 0.8 &  & 0.4          &  & 1.0          & 1.3          & 1.2          &  & \textbf{0.5} &  & 1.2          & 0.8          & 0.7          \\
\textit{Avg} &  & \textbf{0.8} & 1.0 &  & 0.9          &  & 1.7          & 2.3          & 1.9          &  & 0.9          &  & 1.4          & 1.5          & \textbf{1.3} \\
\hline
RC 102-1     &  & 1.6          & 1.6 &  & \textbf{1.4} &  & 1.6          & 1.7          & \textbf{1.4} &  & 1.3          &  & 1.1          & 1.4          & \textbf{1.0} \\
RC 102-2     &  & 0.8          & 1.8 &  & 2.8          &  & 0.8          & 1.9          & \textbf{0.7} &  & 2.4          &  & \textbf{0.3} & 0.4          & 0.8          \\
RC 102-3     &  & 0.8          & 0.8 &  & \textbf{0.6} &  & 0.8          & 0.8          & 1.4          &  & 0.9          &  & 0.5          & \textbf{0.4} & 0.5          \\
RC 102-4     &  & \textbf{0.8} & 1.8 &  & 1.1          &  & 0.9          & 1.7          & 1.5          &  & 1.0          &  & 0.8          & 1.2          & \textbf{0.4} \\
RC 102-5     &  & 1.4          & 1.6 &  & 1.3          &  & \textbf{0.7} & 1.4          & 1.1          &  & 1.8          &  & \textbf{0.7} & 1.0          & 1.1          \\
\textit{Avg} &  & 1.1          & 1.5 &  & 1.4          &  & \textbf{1.0} & 1.5          & 1.2          &  & 1.5          &  & \textbf{0.7} & 0.9          & 0.8          \\
\hline
RC 104-1     &  & 4.8          & 2.4 &  & 3.4          &  & \textbf{0.3} & 0.7          & 0.6          &  & 3.4          &  & \textbf{0.0} & 0.3          & 0.3          \\
RC 104-2     &  & 1.0          & 0.2 &  & 0.3          &  & \textbf{0.0} & \textbf{0.0} & 0.1          &  & 0.3          &  & \textbf{0.0} & \textbf{0.0} & 0.1          \\
RC 104-3     &  & 1.4          & 0.4 &  & 0.3          &  & \textbf{0.0} & \textbf{0.0} & \textbf{0.0} &  & 0.2          &  & \textbf{0.0} & \textbf{0.0} & \textbf{0.0} \\
RC 104-4     &  & 1.6          & 0.2 &  & 0.8          &  & \textbf{0.1} & 0.9          & 0.3          &  & 1.1          &  & \textbf{0.0} & 0.1          & \textbf{0.0} \\
RC 104-5     &  & 2.2          & 0.6 &  & 0.4          &  & \textbf{0.0} & \textbf{0.0} & 0.1          &  & 0.3          &  & 0.1          & \textbf{0.0} & \textbf{0.0} \\
\textit{Avg} &  & 2.2          & 0.8 &  & 1.0          &  & \textbf{0.1} & 0.3          & 0.2          &  & 1.1          &  & \textbf{0.0} & 0.1          & 0.1          \\
\hline
AVG          &  & 1.4          & 1.1 &  & 1.1          &  & \textbf{0.9} & 1.4          & 1.1          &  & 1.1          &  & \textbf{0.7} & 0.8          & \textbf{0.7}
\end{tabular}
\protect\caption{ Detailed results for Class 3 problem instances. Averages over 10 runs. \label{tab:results_class3}}
\end{table}

\begin{table}[p]
\centering
\begin{tabular}{ l  r  c c    r	 c   r   c c c   r   c  r   c c c   }
Class 4 &  &   \multicolumn{8}{c}{10 min. offline computation} &  &  \multicolumn{5}{c}{60 min. offline comput.} \\
         	&   & \multicolumn{2}{c}{MSA}  &     & \multicolumn{1}{c}{GLS}        & 			& \multicolumn{3}{c}{GSA}                  &                   & \multicolumn{1}{c}{GLS}   &      & \multicolumn{3}{c}{GSA}                                                                              \\
  Instance    & \hspace{3em}  & \hspace{0.2em} \textit{d\phantom{fr}} \hspace{0.2em} & \hspace{0.2em} \textit{c\phantom{f}} \hspace{0.2em}  & \hspace{0.7em}  & \hspace{0.2em} \textit{df} \hspace{0.2em}  & \hspace{0.7em}  & \hspace{0.2em} \textit{df} \hspace{0.2em}  & \hspace{0.2em} \textit{dfr} \hspace{0.2em}  & \hspace{0.2em} \textit{ro} \hspace{0.2em} & \hspace{2.5em}  & \hspace{0.2em} \textit{df} \hspace{0.2em} & \hspace{0.7em}  & \hspace{0.2em} \textit{df} \hspace{0.2em}  & \hspace{0.2em} \textit{dfr} \hspace{0.2em}  & \hspace{0.2em} \textit{ro} \hspace{0.2em}  \\
\hline

RC 101-1     &  & \textbf{0.0} & 1.0          &  & 0.3          &  & 1.2          & 1.3          & 1.3          &  & \textbf{0.3} &  & 1.5          & 0.7          & 1.2          \\
RC 101-2     &  & 2.8          & 3.6          &  & 2.3          &  & \textbf{1.9} & 2.1          & 2.6          &  & 2.1          &  & 2.0          & 1.8          & \textbf{1.6} \\
RC 101-3     &  & \textbf{0.0} & 1.6          &  & 0.2          &  & 0.7          & 1.2          & 1.0          &  & 0.5          &  & 0.8          & 0.5          & \textbf{0.2} \\
RC 101-4     &  & 1.0          & 1.4          &  & \textbf{0.7} &  & 2.0          & 2.9          & 1.3          &  & 1.1          &  & \textbf{1.0} & 1.2          & 1.1          \\
RC 101-5     &  & \textbf{2.0} & 2.2          &  & 2.1          &  & 3.0          & 3.9          & 3.8          &  & 1.5          &  & 2.2          & \textbf{1.1} & 3.5          \\
\textit{Avg} &  & 1.2          & 2.0          &  & \textbf{1.1} &  & 1.8          & 2.3          & 2.0          &  & \textbf{1.1} &  & 1.5          & \textbf{1.1} & 1.5          \\
\hline
RC 102-1     &  & \textbf{0.0} & 0.4          &  & 0.1          &  & 0.3          & 0.2          & \textbf{0.0} &  & \textbf{0.1} &  & \textbf{0.1} & \textbf{0.1} & \textbf{0.1} \\
RC 102-2     &  & 1.6          & 1.4          &  & \textbf{0.7} &  & \textbf{0.7} & 1.0          & 0.9          &  & 0.8          &  & 1.2          & \textbf{0.1} & 0.2          \\
RC 102-3     &  & \textbf{0.2} & 1.4          &  & 0.6          &  & 1.7          & 1.5          & 2.1          &  & 1.1          &  & 1.3          & 1.3          & \textbf{0.5} \\
RC 102-4     &  & \textbf{0.0} & \textbf{0.0} &  & 0.1          &  & 0.7          & 0.6          & 0.7          &  & 0.4          &  & 0.2          & \textbf{0.1} & \textbf{0.1} \\
RC 102-5     &  & 0.6          & 0.6          &  & \textbf{0.5} &  & 0.6          & 1.2          & 1.6          &  & \textbf{0.7} &  & 1.0          & 1.0          & 1.5          \\
\textit{Avg} &  & 0.5          & 0.7          &  & \textbf{0.4} &  & 0.8          & 0.9          & 1.1          &  & 0.6          &  & 0.8          & \textbf{0.5} & \textbf{0.5} \\
\hline
RC 104-1     &  & 15.6         & 3.2          &  & 6.7          &  & \textbf{1.4} & 2.0          & 0.9          &  & 5.0          &  & \textbf{0.7} & 0.9          & \textbf{0.7} \\
RC 104-2     &  & 16.0         & 3.4          &  & 4.6          &  & \textbf{0.3} & \textbf{0.3} & 0.4          &  & 4.9          &  & 0.2          & 0.3          & \textbf{0.0} \\
RC 104-3     &  & 13.8         & 5.6          &  & 5.0          &  & \textbf{1.4} & \textbf{1.4} & 1.9          &  & 5.5          &  & 1.3          & 0.9          & \textbf{0.5} \\
RC 104-4     &  & 15.6         & 2.4          &  & 7.3          &  & 0.5          & 2.0          & \textbf{0.2} &  & 6.5          &  & 0.4          & \textbf{0.1} & 0.4          \\
RC 104-5     &  & 8.2          & 2.0          &  & 3.6          &  & 0.5          & 0.5          & \textbf{0.4} &  & 4.1          &  & 1.0          & \textbf{0.2} & 0.7          \\
\textit{Avg} &  & 13.8         & 3.3          &  & 5.4          &  & \textbf{0.8} & 1.2          & \textbf{0.8} &  & 5.2          &  & 0.7          & \textbf{0.5} & \textbf{0.5} \\
\hline
AVG          &  & 5.2          & 2.0          &  & 2.3          &  & \textbf{1.1} & 1.5          & 1.3          &  & 2.3          &  & 1.0          & \textbf{0.7} & 0.8         
\end{tabular}
\protect\caption{ Detailed results for Class 4 problem instances. Averages over 10 runs. \label{tab:results_class4}}
\end{table}

\begin{table}[h]
\centering
\begin{tabular}{ l  r  c c    r	 c c   r   c c    }
Class 6 &  & \multicolumn{2}{p{5.8em}}{No offline computation} &  &  \multicolumn{2}{p{5.8em}}{10 min. offline computation} &  &  \multicolumn{2}{p{5.8em}}{60 min. offline computation} \\
         	&   & GLS & GSA & &  \multicolumn{2}{c}{GSA}  & &  \multicolumn{2}{c}{GSA}  \\
Instance  & \hspace{3em}  &  \hspace{0.2em} \textit{df} \hspace{0.2em} & \hspace{0.2em} \textit{df} \hspace{0.2em}  & \hspace{2.5em} & \hspace{0.2em} \textit{dfr} \hspace{0.2em} & \hspace{0.2em} \textit{ro} \hspace{0.2em} & \hspace{2.5em} & \hspace{0.2em} \textit{dfr} \hspace{0.2em} & \hspace{0.2em} \textit{ro} \hspace{0.2em} \\

\hline

RC 101-1     &  & 6.2          & \textbf{4.8}  &  & \textbf{3.8}  & 4.7           &  & 4.1           & \textbf{3.7}  \\
RC 101-2     &  & 5.7          & \textbf{3.2}  &  & \textbf{2.4}  & 2.8           &  & 3.4           & \textbf{2.5}  \\
RC 101-3     &  & 2.7          & \textbf{2.2}  &  & 4.2           & \textbf{3.6}  &  & \textbf{2.6}  & 3.3           \\
RC 101-4     &  & 6.9          & \textbf{4.9}  &  & 4.4           & \textbf{3.6}  &  & 4.0           & \textbf{3.5}  \\
RC 101-5     &  & \textbf{3.9} & 4.0           &  & 4.4           & \textbf{3.8}  &  & \textbf{3.2}  & 3.7           \\
\textit{Avg} &  & 5.1          & \textbf{3.8}  &  & 3.8           & \textbf{3.7}  &  & 3.5           & \textbf{3.3}  \\
\hline
RC 102-1     &  & 3.7          & \textbf{2.1}  &  & \textbf{3.9}  & 4.2           &  & \textbf{3.1}  & 3.8           \\
RC 102-2     &  & \textbf{2.3} & 3.3           &  & 1.0           & \textbf{0.7}  &  & 1.4           & \textbf{0.6}  \\
RC 102-3     &  & 2.0          & \textbf{0.9}  &  & 1.9           & \textbf{1.4}  &  & \textbf{0.6}  & 0.9           \\
RC 102-4     &  & 2.4          & \textbf{1.8}  &  & 1.4           & \textbf{1.2}  &  & 1.4           & \textbf{1.1}  \\
RC 102-5     &  & 5.4          & \textbf{4.3}  &  & \textbf{3.5}  & 4.2           &  & 4.0           & \textbf{2.6}  \\
\textit{Avg} &  & 3.2          & \textbf{2.5}  &  & \textbf{2.3}  & \textbf{2.3}  &  & 2.1           & \textbf{1.8}  \\
\hline
RC 104-1     &  & 17.6         & \textbf{10.2} &  & \textbf{10.1} & 10.2          &  & 10.8          & \textbf{8.8}  \\
RC 104-2     &  & 15.2         & \textbf{12.4} &  & 13.5          & \textbf{12.0} &  & \textbf{11.7} & 11.9          \\
RC 104-3     &  & 16.6         & \textbf{12.8} &  & 14.5          & \textbf{14.3} &  & 13.2          & \textbf{12.7} \\
RC 104-4     &  & 16.7         & \textbf{16.5} &  & 18.8          & \textbf{17.7} &  & \textbf{16.2} & 17.1          \\
RC 104-5     &  & 16.3         & \textbf{11.3} &  & 14.8          & \textbf{13.8} &  & \textbf{14.2} & 15.9          \\
\textit{Avg} &  & 16.5         & \textbf{12.6} &  & 14.3          & \textbf{13.6} &  & \textbf{13.2} & 13.3          \\
\hline
AVG          &  & 8.2          & \textbf{6.3}  &  & 6.8           & \textbf{6.5}  &  & 6.3           & \textbf{6.1} 
\end{tabular}
\protect\caption{ Detailed results for Class 6 problem instances. Averages over 10 runs. \label{tab:results_class6}}
\end{table}

\newpage


\begin{thebibliography}{10}


\bibitem{ahmed2002sample}
Shabbir Ahmed and Alexander Shapiro.
\newblock {The sample average approximation method for stochastic programs with
  integer recourse}.
\newblock {\em Submitted for publication}, 2002.

\bibitem{asmussen2007stochastic}
S\o~ren Asmussen and Peter~W Glynn.
\newblock {\em {Stochastic Simulation: Algorithms and Analysis: Algorithms and
  Analysis}}, volume~57.
\newblock Springer, 2007.

\bibitem{Bent2004a}
Russell Bent and Pascal~Van Hentenryck.
\newblock {Regrets only! online stochastic optimization under time
  constraints}.
\newblock {\em AAAI}, pages 501--506, 2004.

\bibitem{Bent2004b}
Russell Bent and Pascal~Van Hentenryck.
\newblock {The Value of Consensus in Online Stochastic Scheduling.}
\newblock {\em ICAPS}, (1):219--226, 2004.

\bibitem{Bent2007}
Russell Bent and Pascal~Van Hentenryck.
\newblock {Waiting and Relocation Strategies in Online Stochastic Vehicle
  Routing.}
\newblock {\em IJCAI}, pages 1816--1821, 2007.

\bibitem{Bent2005a}
Russell Bent, Irit Katriel, and Pascal~Van Hentenryck.
\newblock {Sub-optimality approximations}.
\newblock {\em Principles and Practice of Constraint \ldots}, pages 1--15,
  2005.

\bibitem{bent2004scenario}
Russell~W Bent and Pascal {Van Hentenryck}.
\newblock {Scenario-based planning for partially dynamic vehicle routing with
  stochastic customers}.
\newblock {\em Operations Research}, 52(6):977--987, 2004.

\bibitem{Bertsimas1991}
DJ~Bertsimas and G~Van Ryzin.
\newblock {A stochastic and dynamic vehicle routing problem in the Euclidean
  plane}.
\newblock {\em Operations Research}, 1991.

\bibitem{Bertsimas1993}
DJ~Bertsimas and G~Van Ryzin.
\newblock {Stochastic and Dynamic Vehicle Routing in the Euclidean Plane with
  Multiple Capacitated Vehicles}.
\newblock {\em Operations Research}, 1993.

\bibitem{Cordeau2003}
Jean-Fran\c{c}ois Cordeau and Gilbert Laporte.
\newblock {The dial-a-ride problem (DARP): Variants, modeling issues and
  algorithms}.
\newblock {\em 4OR: A Quarterly Journal of Operations Research}, 1(2):89--101,
  2003.

\bibitem{dolan2002benchmarking}
Elizabeth~D Dolan and Jorge~J Mor\'{e}.
\newblock {Benchmarking optimization software with performance profiles}.
\newblock {\em Mathematical programming}, 91(2):201--213, 2002.

\bibitem{flatberg2007dynamic}
Truls Flatberg, Geir Hasle, Oddvar Kloster, Eivind~J Nilssen, and Atle Riise.
\newblock {Dynamic and stochastic vehicle routing in practice}.
\newblock In {\em Dynamic Fleet Management}, pages 41--63. Springer, 2007.

\bibitem{Ghiani2009}
Gianpaolo Ghiani, Emanuele Manni, Antonella Quaranta, and Chefi Triki.
\newblock {Anticipatory algorithms for same-day courier dispatching}.
\newblock {\em Transportation Research Part E: Logistics and Transportation
  Review}, 45(1):96--106, January 2009.

\bibitem{Hvattum2006}
Lars~M. Hvattum, Arne L\o~kketangen, and Gilbert Laporte.
\newblock {Solving a Dynamic and Stochastic Vehicle Routing Problem with a
  Sample Scenario Hedging Heuristic}.
\newblock {\em Transportation Science}, 40(4):421--438, November 2006.

\bibitem{Ichoua2006}
Soumia Ichoua, Michel Gendreau, and Jean-Yves Potvin.
\newblock {Exploiting Knowledge About Future Demands for Real-Time Vehicle
  Dispatching}.
\newblock {\em Transportation Science}, 40(2):211--225, May 2006.

\bibitem{kindervater1997vehicle}
Gerard A~P Kindervater and Martin W~P Savelsbergh.
\newblock {Vehicle routing: handling edge exchanges}.
\newblock {\em Local search in combinatorial optimization}, pages 337--360,
  1997.

\bibitem{Mitrovic-Minic2004a}
Sne\v{z}ana Mitrovi\'{c}-Mini\'{c} and Gilbert Laporte.
\newblock {Waiting strategies for the dynamic pickup and delivery problem with
  time windows}.
\newblock {\em Transportation Research Part B: Methodological}, 38(7):635--655,
  August 2004.

\bibitem{pillac2013review}
Victor Pillac, Michel Gendreau, Christelle Gu\'{e}ret, and Andr\'{e}s~L
  Medaglia.
\newblock {A review of dynamic vehicle routing problems}.
\newblock {\em European Journal of Operational Research}, 225(1):1--11, 2013.

\bibitem{Pillac2012}
Victor Pillac, Christelle Gu\'{e}ret, and Andr\'{e}s~L. Medaglia.
\newblock {An event-driven optimization framework for dynamic vehicle routing}.
\newblock {\em Decision Support Systems}, 54(1):414--423, December 2012.

\bibitem{psaraftis1980dynamic}
Harilaos~N Psaraftis.
\newblock {A dynamic programming solution to the single vehicle many-to-many
  immediate request dial-a-ride problem}.
\newblock {\em Transportation Science}, 14(2):130--154, 1980.

\bibitem{psaraftis1995dynamic}
Harilaos~N Psaraftis.
\newblock {Dynamic vehicle routing: Status and prospects}.
\newblock {\em annals of Operations Research}, 61(1):143--164, 1995.

\bibitem{Schilde2011}
M~Schilde, K~F Doerner, and R~F Hartl.
\newblock {Metaheuristics for the dynamic stochastic dial-a-ride problem with
  expected return transports.}
\newblock {\em Computers \& operations research}, 38(12):1719--1730, December
  2011.

\bibitem{shapiro2009lectures}
Alexander Shapiro, Darinka Dentcheva, and Andrzej~P Ruszczy$\backslash$'nski.
\newblock {\em {Lectures on stochastic programming: modeling and theory}},
  volume~9.
\newblock SIAM, 2009.

\bibitem{shaw1998using}
Paul Shaw.
\newblock {Using constraint programming and local search methods to solve
  vehicle routing problems}.
\newblock In {\em Principles and Practice of Constraint Programming—CP98},
  pages 417--431. Springer, 1998.

\bibitem{Solomon1987}
MM~Solomon.
\newblock {Algorithms for the vehicle routing and scheduling problems with time
  window constraints}.
\newblock {\em Operations research}, 35(2), 1987.

\bibitem{Taillard1997}
\'{E} Taillard and P~Badeau.
\newblock {A tabu search heuristic for the vehicle routing problem with soft
  time windows}.
\newblock {\em Transportation \ldots}, pages 1--36, 1997.

\bibitem{VanHentenryck2009}
Pascal {Van Hentenryck}, Russell Bent, and Eli Upfal.
\newblock {\em {Online stochastic optimization under time constraints}}, volume
  177.
\newblock September 2009.

\bibitem{verweij2003sample}
Bram Verweij, Shabbir Ahmed, Anton~J Kleywegt, George Nemhauser, and Alexander
  Shapiro.
\newblock {The sample average approximation method applied to stochastic
  routing problems: a computational study}.
\newblock {\em Computational Optimization and Applications}, 24(2-3):289--333,
  2003.

\bibitem{wilson1977computer}
Nigel H~M Wilson and Neil~J Colvin.
\newblock {\em {Computer control of the Rochester dial-a-ride system}}.
\newblock Massachusetts Institute of Technology, Center for Transportation
  Studies, 1977.


\bibitem{saintguillain2015multistage}
Michael Saint-Guillain, Yves Deville and Christine Solnon.
\newblock {A Multistage Stochastic Programming Approach to the Dynamic and Stochastic VRPTW}.
\newblock {\em Submitted for publication}, 2015.

\end{thebibliography}

\end{document}